# Article information

**Article title:** BDSL 49: A Comprehensive Dataset of Bangla Sign Language.

## Authors

Ayman Hasib[a,†], Saqib Sizan Khan[a,†], Jannatul Ferdous Eva[a,†], Mst. Nipa Khatun[a,‡], Ashraful Haque[a,‡], Nishat Shahrin[a,‡], Rashik Rahman[a,‡], Hasan Murad[b], Md. Rajibul Islam[c,*], Molla Rashied Hussein[a,*]

## Affiliations:

[a] Department of Computer Science and Engineering, University of Asia Pacific, Dhaka, Bangladesh.

[b] Department of Computer Science and Engineering, Chittagong University of Engineering & Technology, Chittagong, Bangladesh.

[c] Department of Computer Science and Engineering, Bangladesh University of Business and Technology, Dhaka, Bangladesh.

[*]**The correspondent author's email address and Twitter handle**
**Email address:** md.rajibul.islam@bubt.edu.bd (M.R. Islam), mrh.cse@uap-bd.edu (M.R. Hussein)
[†]Equal contribution and joint first author.
[‡]Equal contribution and joint second author.

## Abstract
Language is a method by which individuals express their thoughts. Each language has its own set of alphabetic and numeric characters. People can communicate with one another through either oral or written communication. However, each language has a sign language counterpart. Individuals who are deaf and/or mute communicate through sign language. The Bangla language also has a sign language, which is called BDSL. The dataset is about Bangla hand sign images. The collection contains 49 individual Bangla alphabet images in sign language. BDSL49 is a dataset that consists of 29,490 images with 49 labels. Images of 14 different adult individuals, each with a distinct background and appearance, have been recorded during data collection. Several strategies have been used to eliminate noise from datasets during preparation. This dataset is available to researchers for free. They can develop automated systems using machine learning, computer vision, and deep learning techniques. In addition, two models were used in this dataset. The first is for detection, while the second is for recognition.

## Specifications Table:

| | |
|---|---|
| **Subject** | Sign Language, Deep Learning, Computer Vision, Natural Language Processing |
| **Specific subject area** | Individuals who have difficulty in hearing communicate using sign language. Images of Bangla sign language have been collected to convert into Bangla alphabet to improve communication. Techniques such as computer vision, machine learning, deep learning, and natural language processing have been implemented. |
| **Type of data** | The detection dataset is in full-frame images. The recognition dataset contains cropped hand sign images, which are in JPG format. |
| **How the data were acquired** | Smartphone cameras were used to capture hand gestures corresponding to hand signs. |
| **Data format** | Fully labelled and raw RGB images. |
| **Description of data collection** | 49 distinct categories in this dataset are labelled with the Bangla alphabet. Each category contains approximately 600 images to culminate into 29,490 images. |
| **Data source locations** | Department of Computer Science and Engineering, University of Asia Pacific, Dhaka, Bangladesh. |

| **Data accessibility** | **Repository name:** BDSL 49: A Comprehensive Dataset of Bangla Sign Language<br><br>**Data identification number:** 10.17632/k5yk4j8z8s.5<br><br>**Direct URL to data:** https://data.mendeley.com/datasets/k5yk4j8z8s/5<br><br>**Instructions for accessing this dataset:** This dataset is publicly available for any research, academic, or educational purpose at the repository hyperlink mentioned above. |
|---|---|
| **Related research articles** | Md. Sanzidul Islam, S. Sultana Sharmin Mousumi, N.A. Jessan, A. Shahariar Azad Rabby, S. Akhter Hossain, Ishara-Lipi: The First Complete MultipurposeOpen Access Dataset of Isolated Characters for Bangla Sign Language, in: 2018 International Conference on Bangla Speech and Language Processing (ICBSLP), IEEE, 2018. http://dx.doi.org/10.1109/icbslp.2018.8554466. |

**Value of the data**

- This dataset, which is based on Bangla hand signs, will be extremely useful as a means of communication for those who are deaf and/or mute.
- This dataset will benefit the general public who are not part of the deaf and/or mute populations. It will also be valuable for non-disabled individuals who are unfamiliar with sign language or the specific sign(s), and it may assist to reduce the inequity between the deaf and/or mute populations and non-disabled communities.
- This dataset can be used for training in expert systems. The suggested sign language dataset opens up a plethora of new avenues for future study and development. It offers potential advantages in machine learning and artificial intelligence applications.
- The proposed dataset can be used to train a model for smartphone applications such as capturing hand sign image in real-time or live video streaming.

## Data description

About 3 million individuals in Bangladesh are deaf or hearing-impaired [1]. Sign language is a communication technique by which deaf or mute people can communicate with non-disabled people via hand gestures. Verbal communication is quite difficult for deaf and/or mute people. Moreover, they face a significant challenge because non-disabled people fail to understand these complicated hand signs. For this reason, a hand sign dataset based on the Bangla language has been created which represents the Bangla alphabet and numeric digits. The Bangla Sign Language dataset was created to bridge the gap between the deaf and/or mute people and the non-disabled communities. This dataset is one of several available in Bangla Sign Language. It helps to improve overall accuracy in sign detection and identification. It has been made publicly available to help and encourage more research in Bangla Sign Language Interpretation so that hearing and speech impaired people can be benefited from it. Most of the previous work on the BDSL dataset deals with around 30–38 labels [2–5], but this dataset deals with 49 labels that have not been applied anywhere else. After acquiring 14,745 images from a few individuals, two datasets have been produced from these images. One is an annotated detection dataset (full-frame image). The second is the recognition dataset (cropped image). This processing of a single image is adapted by [6] and YOLOv4 methodologies were adapted from [7, 8].

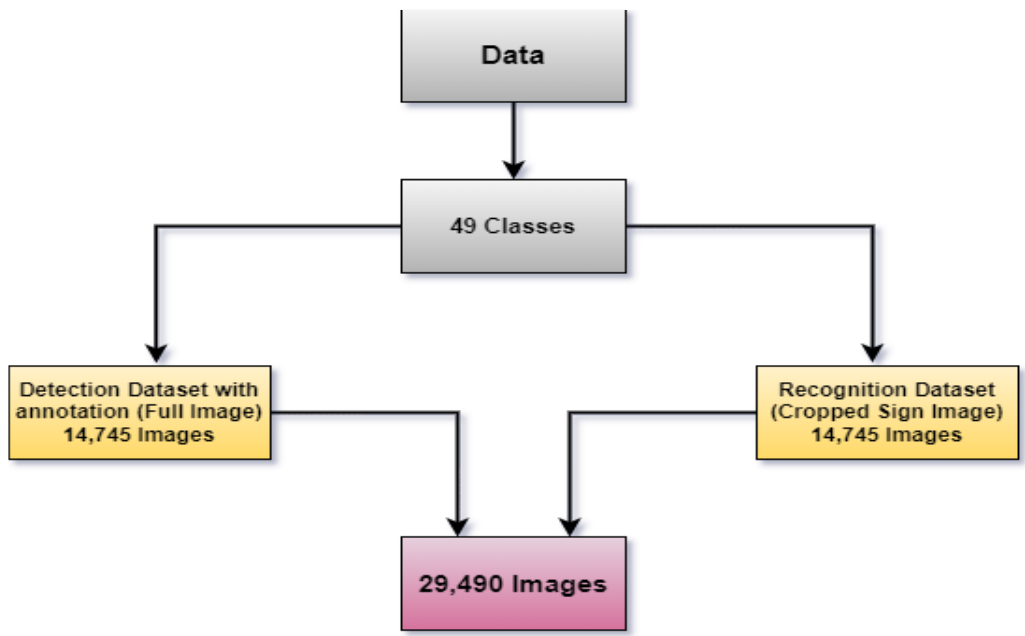

**Figure 1: The Dataset Creation Process**

This dataset comprises around 37 alphabets, 10 numeric characters, and 2 special characters to properly recognize Bangla symbols. The dataset consists of 29,490 images divided into two sections, where each section is further divided into 49 classes as shown in Figure 1. In this dataset, alphabetic and numeric characters are classified based on the shape and orientation of the hand signals, as seen in Figures 2 and 3. The first section contains full-frame images from the detection dataset, whereas the second contains solely cropped images of the hand signs from the recognition dataset. In this dataset, around 300 photos from each class are collected using various smartphone cameras, totalling 14,745 images. Following that, these images are created by cutting the hand sign region of the full-frame image from the detection dataset. Both the detection and recognition datasets are divided into training and testing datasets, with the training dataset including 80% of the total data and the testing dataset containing 20%. Figure 2.1 to 2.3 illustrate some samples of the detection dataset, whereas Figure 3.1 and 3.2 depict the recognition dataset. Data distribution of images among each class is shown in Table 1 where the total number of images is an aggregation of both sections.

**Table 1:** Image distribution among classes considering both sections of dataset

| Label No. | Class Name | Total Number of images | Label No. | Class Name | Total Number of images |
|---|---|---|---|---|---|
| 1 | অ | 598 | 26 | ভ | 600 |
| 2 | আ | 638 | 27 | ম | 600 |
| 3 | ই | 592 | 28 | য় | 600 |
| 4 | উ | 600 | 29 | র | 600 |
| 5 | এ | 600 | 30 | ল | 600 |
| 6 | ও | 600 | 31 | ন | 600 |
| 7 | ক | 600 | 32 | স | 600 |
| 8 | খ | 600 | 33 | হ | 600 |
| 9 | গ | 602 | 34 | ড় | 598 |
| 10 | ঘ | 600 | 35 | ২ং | 600 |
| 11 | চ | 600 | 36 | ২ঃ | 600 |

| 12 | ছ | 600 | 37 | ০ | 600 |
| --- | --- | --- | --- | --- | --- |
| 13 | জ | 600 | 38 | ১ | 600 |
| 14 | ঝ | 600 | 39 | ২ | 600 |
| 15 | ট | 600 | 40 | ৩ | 600 |
| 16 | ঠ | 600 | 41 | ৪ | 600 |
| 17 | ড | 600 | 42 | ৫ | 596 |
| 18 | ঢ | 600 | 43 | ৬ | 600 |
| 19 | ত | 600 | 44 | ৭ | 600 |
| 20 | থ | 600 | 45 | ৮ | 600 |
| 21 | দ | 600 | 46 | ৯ | 604 |
| 22 | ধ | 600 | 47 | ◌ং | 600 |
| 23 | প | 604 | 48 | space | 600 |
| 24 | ফ | 596 | 49 | ঞ | 600 |
| 25 | ব | 600 | | | |

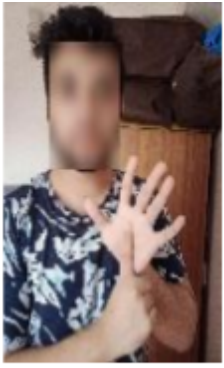 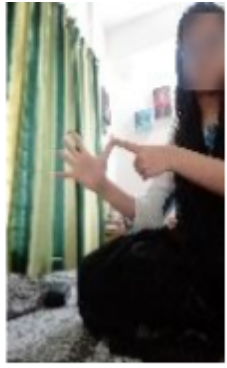 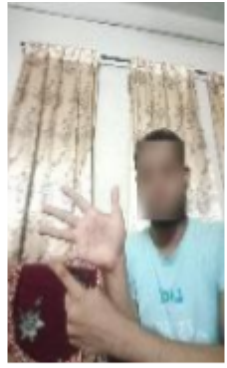 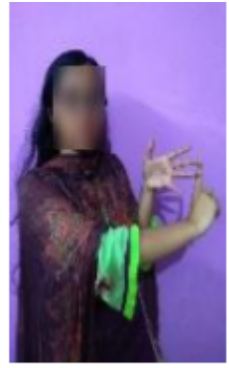 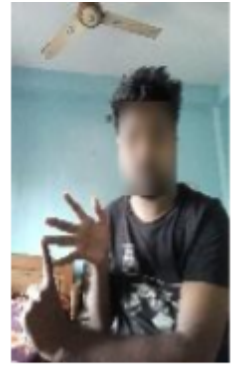

অ   আ   ই   উ   এ

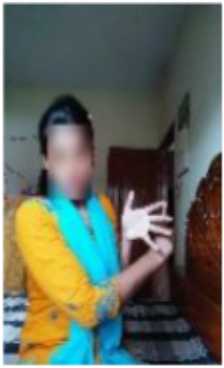 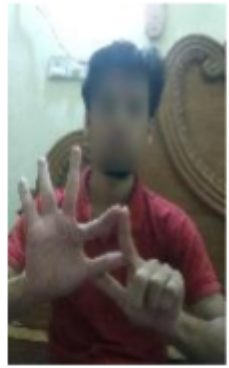 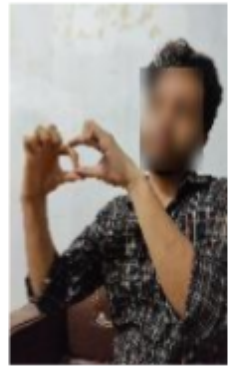 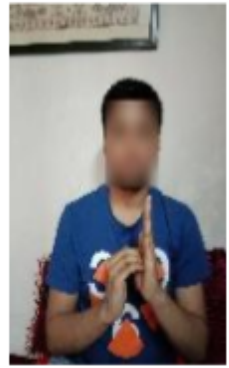 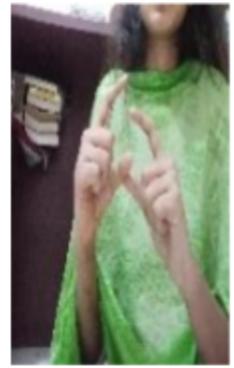

ও   ক   খ   গ   ঘ

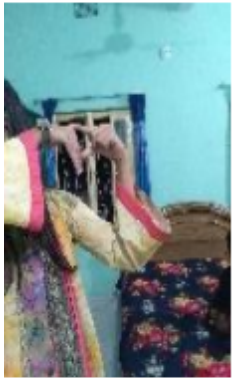 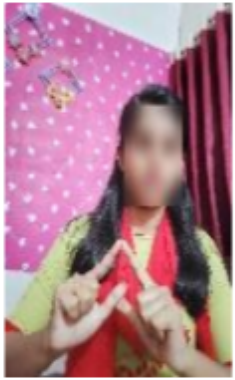 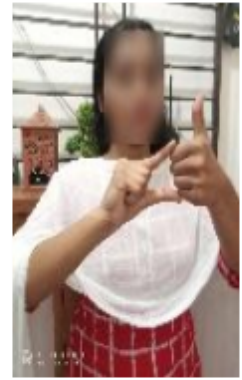 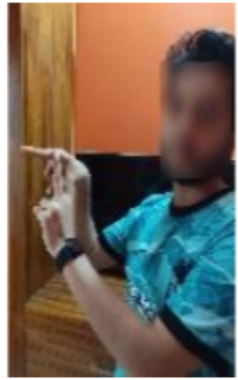 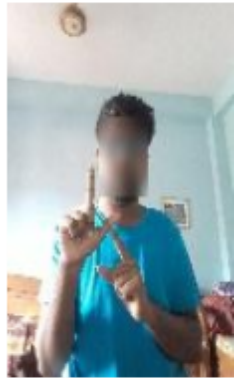

চ   ছ   জ   ঝ   ট

**Figure 2.1: Detection dataset (Full-frame) from labels 1-15**

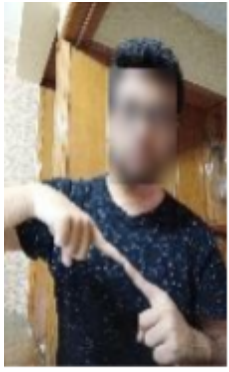 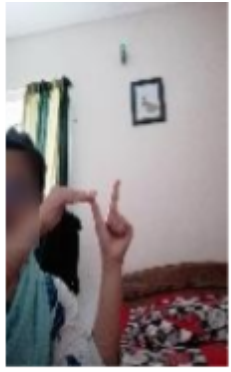 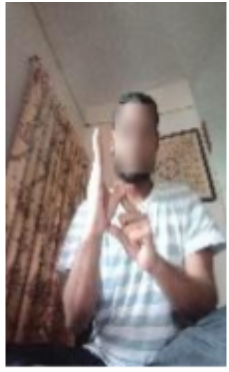 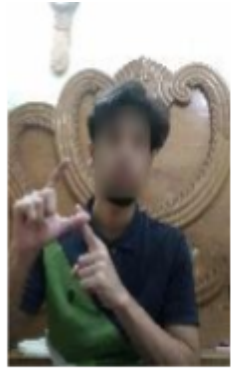 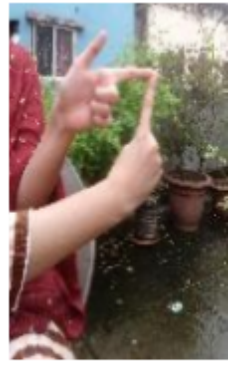

ঠ    ড    ঢ    ত    থ

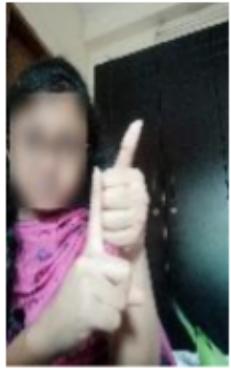 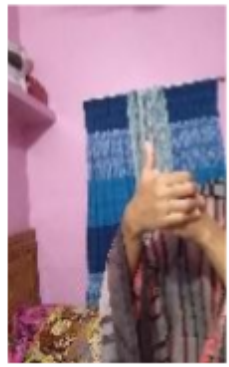 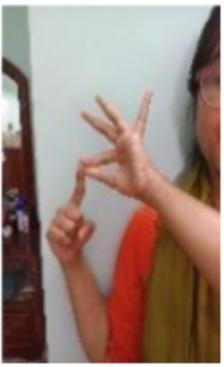 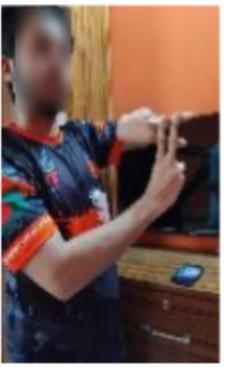 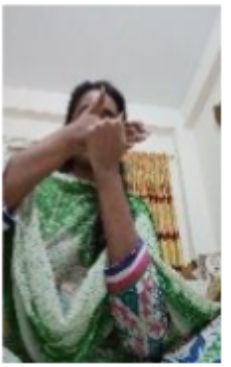

দ    ধ    প    ফ    ব

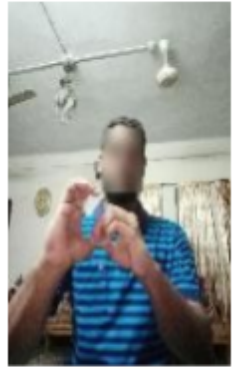 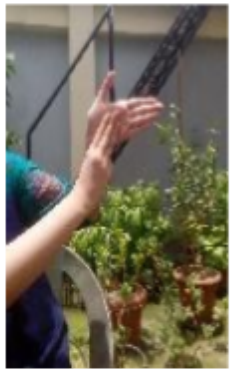 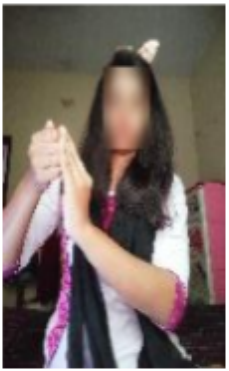 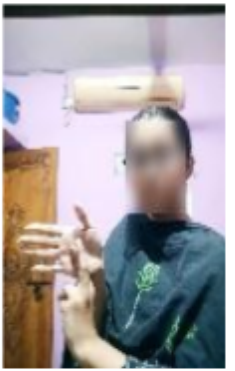 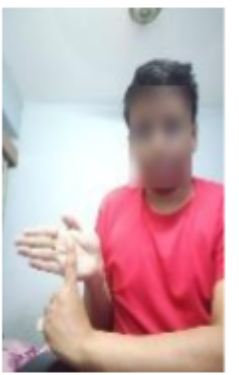

ভ    ম    য    র    ল

**Figure 2.2: Detection dataset (Full-frame) from labels 16-30**

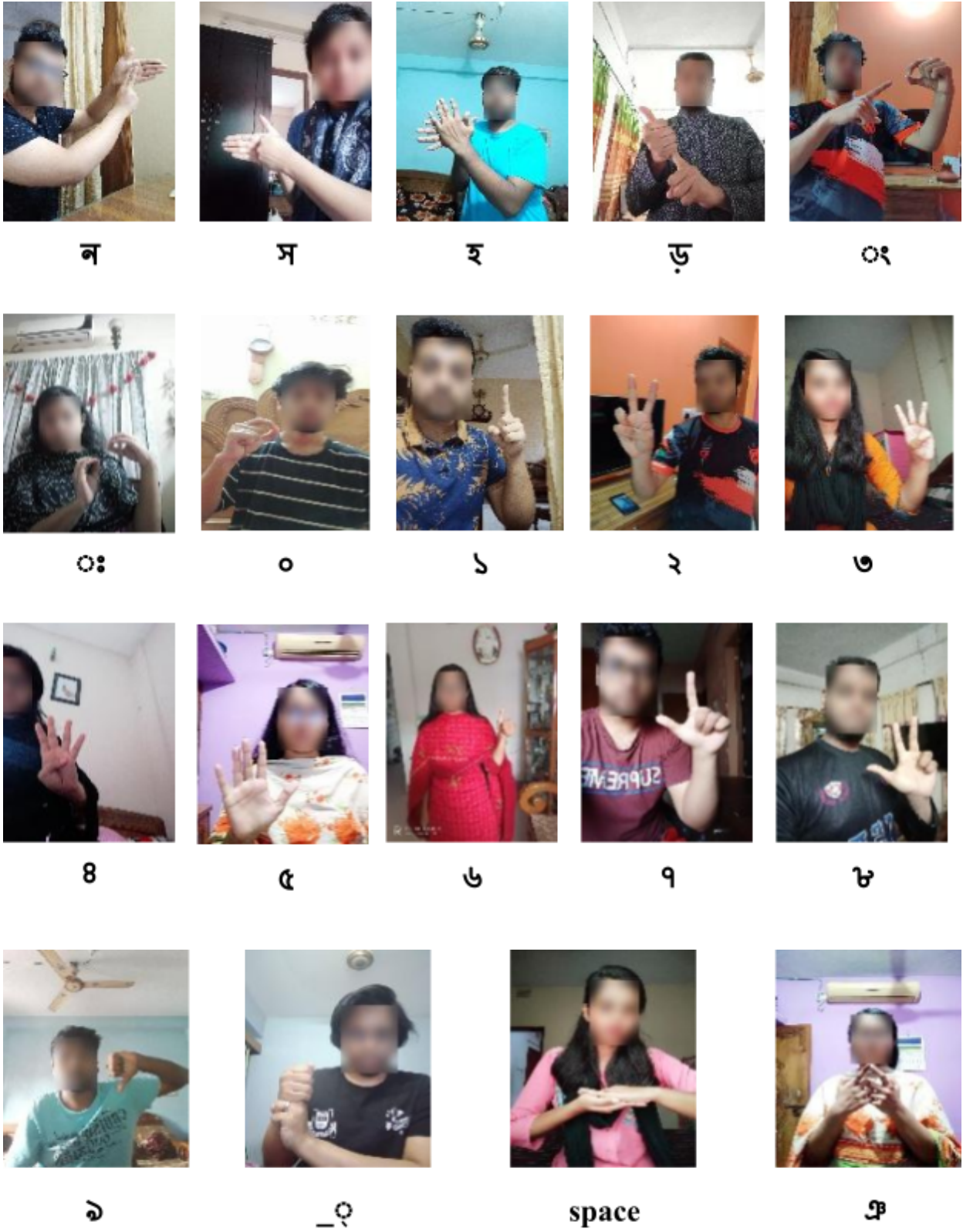

**Figure 2.3: Detection dataset (Full-frame) from labels 31-49**

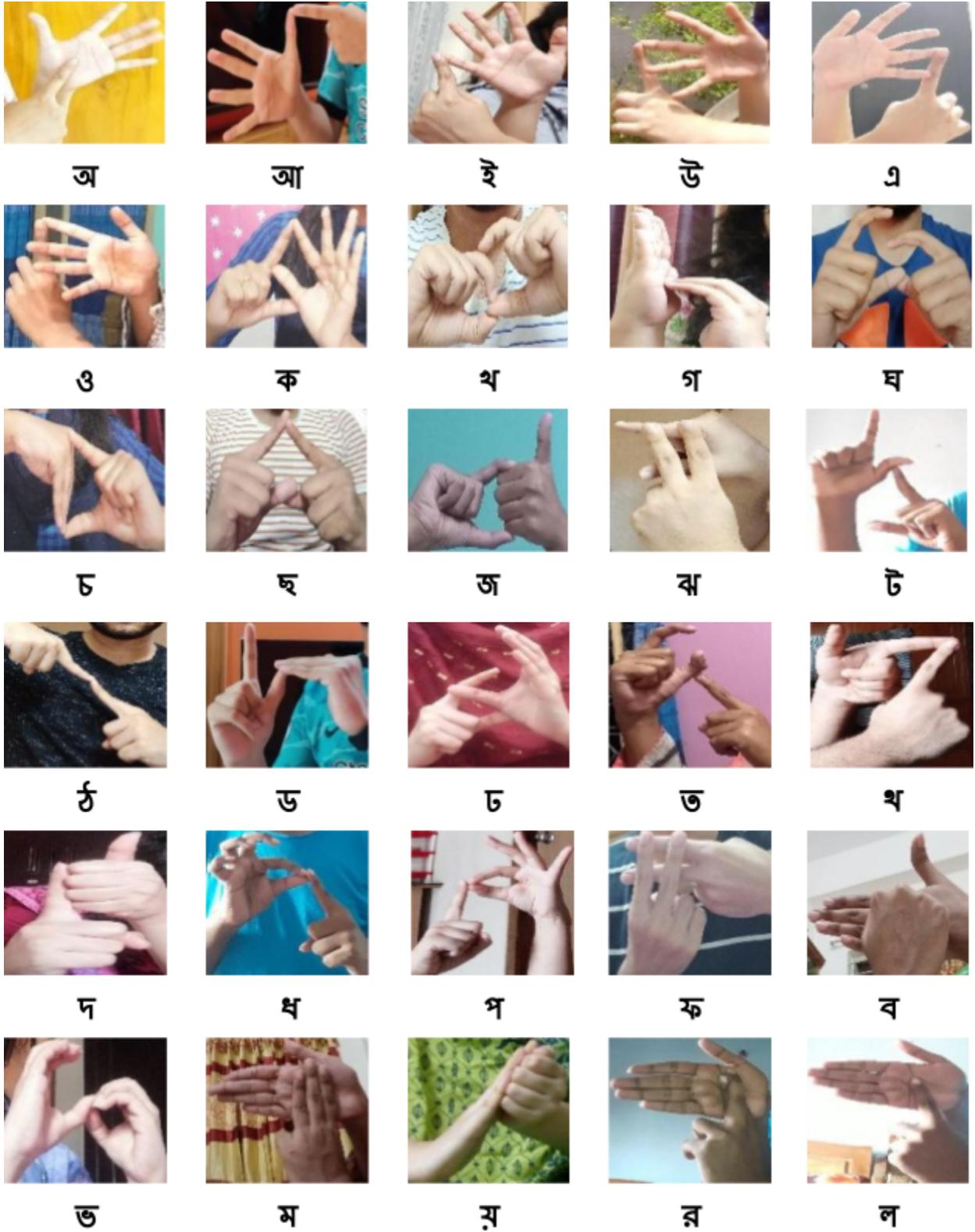

Figure 3.1: Recognition dataset (Cropped Images) from labels 1-30

**Figure 3.2: Recognition dataset (Cropped Images) from labels 31-49**

Figures 2.1, 2.2, and 2.3 show the sample dataset images in full-frame. This section of dataset images is useful for the Detection model. Correspondingly, Figures 3.1 and 3.2 depict the sample dataset images of hand signs that have been extracted from the full-frame images. These cropped images are useful for the Recognition model.

**Experimental design, materials, and methods**

This dataset has two sections for creating two distinct types of models, one for detection and one for recognition. We utilized full-frame RGB photographs with varied backgrounds, as shown in Figure 4, and afterward annotated them to highlight where the sign appeared. As illustrated in this figure, all the faces in the dataset images were detected and blurred using Python and the OpenCV package.

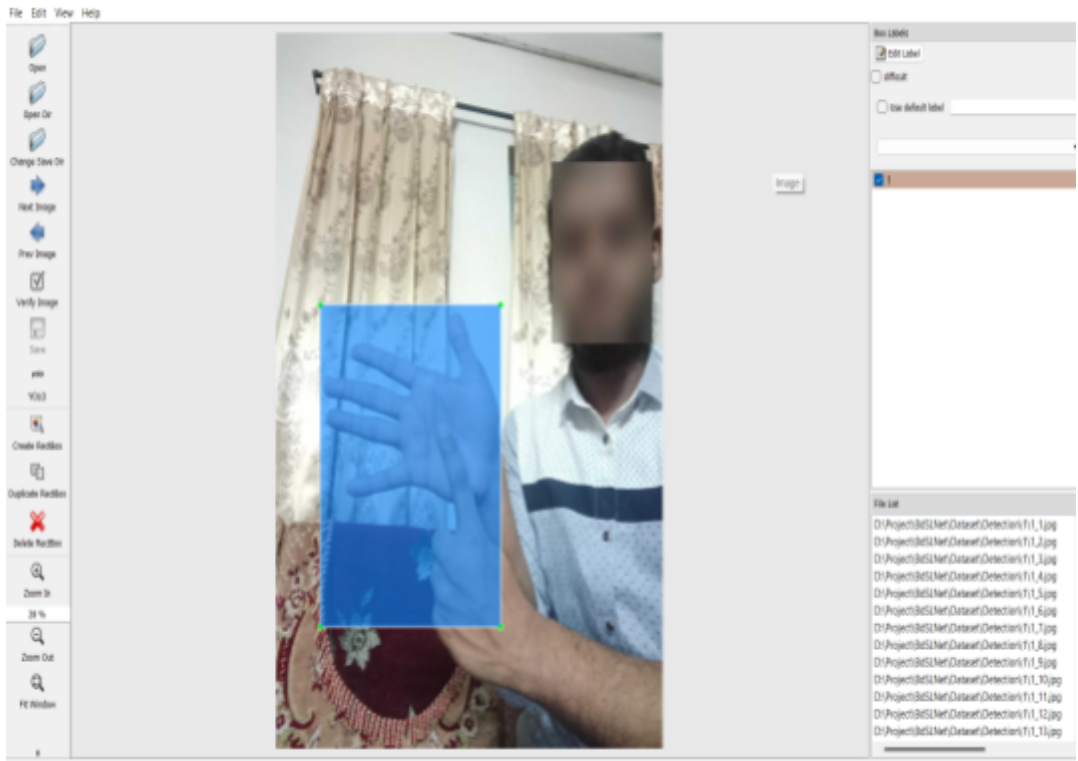

**Figure 4: Annotation Process**

Deep learning models were employed in both parts to evaluate model development potential with this dataset. A real-time detection model named YOLOV4 was used for the detection part, and the model attained an accuracy of 99.4 percent. For recognition, pre-trained models such as Xception, InceptionV3, InceptionResNetv2, ResNet50V2, and DenseNet were employed. Figure 5 shows accuracy comparisons for all of the models that were trained and assessed on the same dataset.

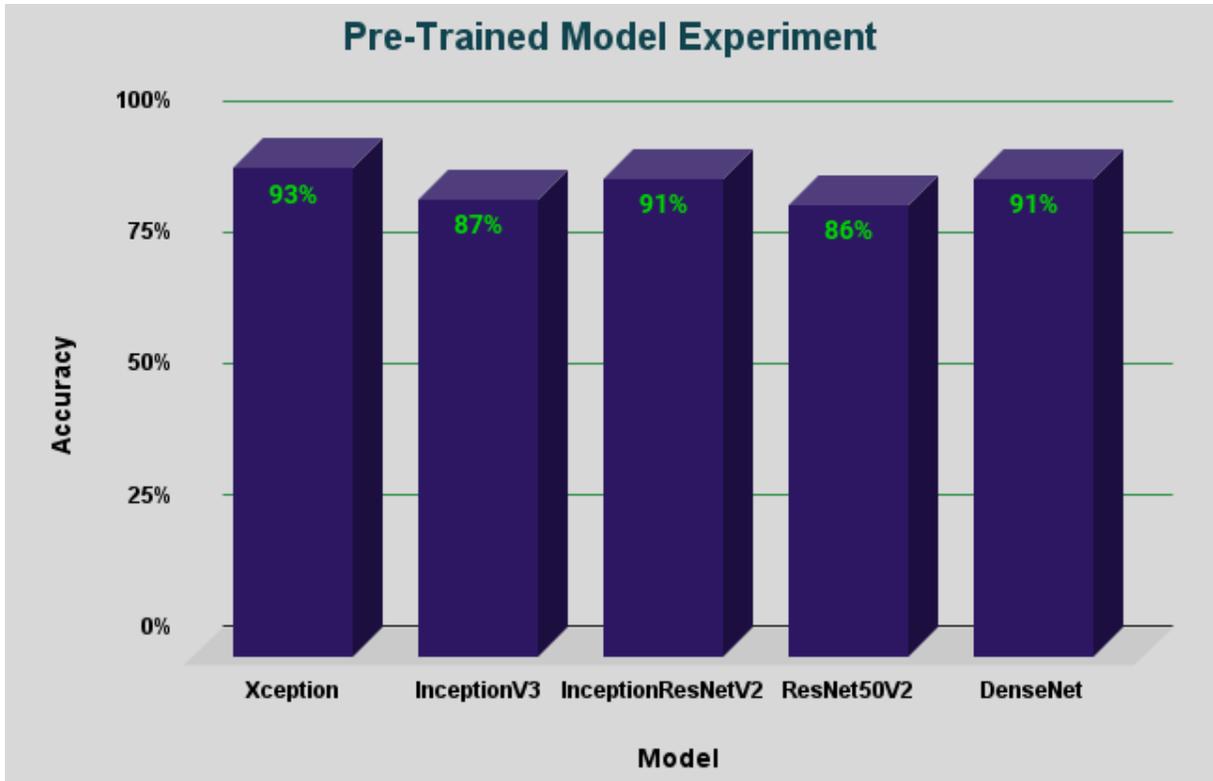

**Figure 5: Accuracy comparison of various pre-trained models on the recognition dataset.**

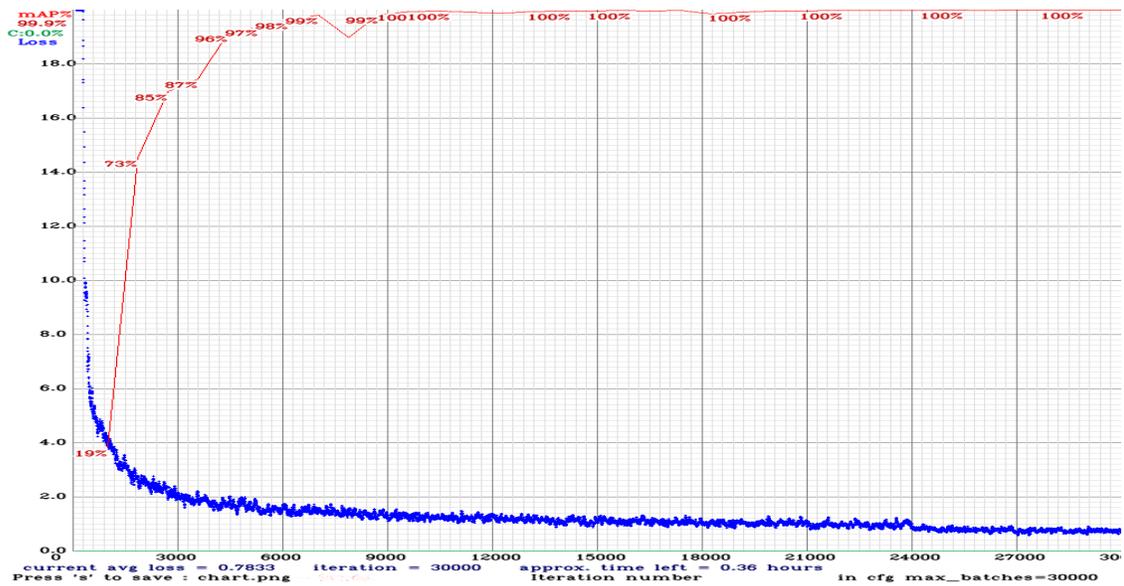

**Figure 6: Mean-average precision chart of the YOLOv4 model on the detection dataset**

Here we have implemented various pre-trained models on the recognition dataset and they achieved 93% accuracy in Xception, 87% accuracy in InceptionV3, 91% InceptionResNetV2, 86% accuracy in ResNet50V2, and 91% accuracy in DenseNet.
From this figure, it is to be noted that the Xception model gets the highest accuracy in recognizing the sign languages.

On the other hand, Figure 6 illustrates the mean average accuracy and loss curve on the YOLOV4 detection model, which detects sign languages from images with a mean average precision of 99.9%. It is a highly accurate and generalized detection model.

Before training the YOLOv4 model, the data must be annotated. Each 'img-name.png' image is accompanied by an 'img name.txt' file that contains annotation information. The text file includes the image's class number, x min coordinate, x-max coordinate, y-min coordinate, and y-max coordinates. Python and a couple of packages, such as lxml and pyqt5, are required to use this utility. After installing these packages, we launch the application and load images from the specified folder. Then, a bounding box is manually drawn around the object to be detected. The annotation format must be YOLO in order to be able to annotate the images. The bounding box is a precisely drawn rectangle surrounding the detecting zone that is subsequently assigned a class. The coordinates of this box are determined after sketching the region. These coordinates must then be normalized depending on the image's height and width before being saved as the annotated document in the previously stated text format. The annotating procedure follows algorithm 1.

**Algorithm 1:**

*ImgList = getAllImginDir(path)*
***while*** *len(Img List)≠ 0 do*
    *img = ImgList.pop()*
    *class_no = getClassNo()*
    *x, y, width, height = gelCoordinates(Draw(img))*
    *x,y,width,height=(x/width),(y/height),(width/img_width),(height/img_height)*
    *write(class_no, x, y, width, height) as txt*
***end while***

**Ethics statements**

In this dataset, all contributors voluntarily participated in the dataset's creation. Smartphone cameras captured all the hand-sign images. Using cameras, we only captured the hand gestures of the volunteers. In that case, there is no impact on our bodies. Therefore, there are no health-related ethical issues.

**Credit author statement**

**Ayman Hasib:** Validation, Investigation, Data Curation, Writing - Original Draft, Visualization. **Saqib Sizan Khan:** Methodology, Validation, Investigation, Data Curation, Writing-Original Draft, Visualization. **Jannatul Ferdous Eva:** Validation, Investigation, Data Curation, Writing - Original Draft, Visualization. **Rashik Rahman:** Writing-Review & Editing, Supervision, Validation, Project Administration, **Nipa Khatun:** Data Curation, Writing - Original Draft, **Ashraful Haque:** Data Curation, Writing - Original Draft, **Nishat Shahrin:** Data Curation, Writing - Original Draft **Hasan Murad:** Supervision, Project Administration **Md. Rajibul Islam:** Writing-Review & Editing, Supervision, Project Administration. **Molla Rashied Hussein:** Writing-Review & Editing, Supervision.

**Acknowledgments**

We would like to express our appreciation and gratitude to the following members of the [data collection team](): **Fariha Nusrat, Rashed Khandoker, Jolekha Begum Bristy, Kashfia Jashim, Sabrina Jahan, Azra Tuni, Abdullah Mohammad Sakib, Nakiba Nuren Rahman,** and **Md. Imran Nazir**. They all participate in capturing the sign images. They only work as volunteers to collect data.

**Declaration of interests**

■ The authors declare that they have no known competing financial interests or personal relationships that could have appeared to influence the work reported in this paper.

☐ The authors declare the following financial interests or personal relationships which may be considered as potential competing interests: